\documentclass[runningheads]{llncs}

\usepackage{eccv}
\usepackage[symbol]{footmisc}

\usepackage{eccvabbrv}

\usepackage{graphicx}
\usepackage{booktabs}
\usepackage{mathrsfs}
\usepackage{multirow}
\usepackage[accsupp]{axessibility}  

\usepackage{hyperref}

\usepackage{orcidlink}
\usepackage[linesnumbered,ruled,vlined]{algorithm2e}

\begin{document}

\title{Diff-Tracker: Text-to-Image Diffusion Models are Unsupervised Trackers}

\titlerunning{Diff-Tracker: Text-to-Image Diffusion Models are Unsupervised Trackers}

\author{Zhengbo Zhang\inst{1}\orcidlink{0009-0000-8956-0095} \and
Li Xu\inst{1}\orcidlink{0000-0003-1575-5724} \and
Duo Peng\inst{1}\orcidlink{0000-0003-3281-0772} \and \\
Hossein Rahmani \inst{2}\orcidlink{0000-0003-1920-0371} \and
Jun Liu  \inst{1,2} \thanks{corresponding author}\orcidlink{0000-0002-4365-4165}}

\authorrunning{Zhengbo Zhang, Li Xu et al.}

 \institute{Singapore University of Technology and Design \\
\email{\{zhengbo\_zhang, li\_xu, duo\_peng\}@mymail.sutd.edu.sg}\and 
Lancaster University \\
 \email{h.rahmani@lancaster.ac.uk}, \email{j.liu81@lancaster.ac.uk}}

\maketitle

\begin{abstract}
We introduce Diff-Tracker, a novel approach for the challenging unsupervised visual tracking task leveraging the pre-trained text-to-image diffusion model. Our main idea is to leverage the rich knowledge encapsulated within the pre-trained diffusion model, such as the understanding of image semantics and structural information, to address unsupervised visual tracking. To this end, we design an initial prompt learner to enable the diffusion model to recognize the tracking target by learning a prompt representing the target. Furthermore, to facilitate dynamic adaptation of the prompt to the target's movements, we propose an online prompt updater. Extensive experiments on five benchmark datasets demonstrate the effectiveness of our proposed method, which also achieves state-of-the-art performance.
\end{abstract}

\section{Introduction}
\label{sec:intro}

Visual object tracking constitutes a core task in the field of computer vision, finding extensive applications in areas ranging from autonomous driving~\cite{chen2021novel,gao2019manifold} to robotics~\cite{papanikolopoulos1993visual,budiharto2020design}. In the recent development in visual object tracking, deep learning based trackers~\cite{yu2020deformable,danelljan2020probabilistic,yan2021learning,ye2022joint,lin2022swintrack,cui2022mixformer,mayer2022transforming,yuan2020self} have emerged as a prevailing paradigm. These trackers exhibit a strong reliance on data, necessitating a substantial volume of annotated data for supervised training. Owing to the high cost and time demands associated with manual data annotation, unsupervised visual tracking has recently experienced a surge in attention.  Although previous researchers have made significant efforts~\cite{zheng2021learning,shen2022unsupervised,wang2021unsupervised}, unsupervised object tracking remains a substantial challenge due to the difficulty in effectively exploiting the rich semantic and structural information of video frames~\cite{wang2021unsupervised,sio2020s2siamfc}, as well as the abundant contextual relationships within videos~\cite{zheng2021learning}.

At the same time, the pre-trained text-to-image diffusion model has achieved exemplary performance across various realms, including text-to-image generation~\cite{rombach2022high,hertz2022prompt} and text-to-video generation~\cite{khachatryan2023text2video}. For instance, in the text-to-image area, the pre-trained text-to-image diffusion model, such as Stable Diffusion~\cite{rombach2022high}, has demonstrated remarkable capabilities in generating images that are not only diverse but also rich in detail and adhere to reasonable spatial structures, all in response to user-defined prompts. Such impressive results imply that the pre-trained text-to-image diffusion model exhibits an extensive understanding of visual representations, ranging from pixel-level semantic details to spatial layouts, including aspects like object texture, object shape, and spatial arrangement within the image. Additionally, researchers~\cite{tang2023emergent} verify that the pre-trained diffusion model, despite not being trained or fine-tuned on video data, exhibits commendable capabilities in understanding video contextual relationships.  Given that the pre-trained text-to-image diffusion model demonstrates the understanding (knowledge) of the semantic and structural information of video frames, and contextual relationships within video, a natural question arises: \textit{Can we leverage the rich knowledge encapsulated in the pre-trained text-to-image diffusion model for performing unsupervised visual tracking?}

Nevertheless, it is non-trivial to leverage the knowledge implicitly embedded in the pre-trained text-to-image diffusion model for unsupervised object tracking in videos. Such models are engineered for image generation from text prompts and are not inherently suited for visual object tracking. To handle this problem, we delve into the underlying characteristic of the text-to-image diffusion model. \textbf{Firstly}, we revisit the input type (text prompts) and output type (images) of the text-to-image diffusion model, interpreting the model's function from a novel perspective. We view the text-to-image diffusion model as a bridge that connects the semantic information of the input prompts with the content of the output images. \textbf{Secondly}, this semantic connection between text prompts and output images is specifically manifested in the cross-attention maps within the text-to-image diffusion model~\cite{khani2023slime}. The text prompts are capable of activating areas on the cross-attention maps that are semantically related to the prompts. In this paper, activated regions mean that the pixel values in these areas on the attention maps are high. This suggests that if we learn a prompt for the tracking target, the prompt can enable the text-to-image diffusion model to activate region of the target object on the cross-attention maps of input video frames. Consequently, we can leverage the rich knowledge embedded within the pre-trained text-to-image diffusion model to perform the challenging unsupervised object tracking.

However, learning a prompt for the target object is not straightforward: i) Since the relationship between the target and the background is beneficial for tracking the target in scenarios where it undergoes significant appearance deformations or occlusions~\cite{zhu2015weighted,fang2017spatial}, it presents a challenge that the learned prompt should encode rich relationships between the target object and backgrounds.  ii) The continuous motion of the target object leads to changes in the appearance of the object, as well as the relationships between the target object and the background. This poses another challenge of how to update the prompt online to adapt to these changes. 

To address these challenges, we propose Diff-Tracker, which consists of an initial prompt learner and an online prompt updater. The initial prompt learner is devised to learn an initial prompt for tracking the target object. To be exact, through this module, we learn the initial prompt that can accurately activate the target object's region in the cross-attention map of the first frame.  To encapsulate more information about the relationship between the target object and the background, we design an attention harmonization method to combine the original cross-attention map and cross-attention map enhanced by the self-attention map from the text-to-image diffusion model. 

The designed online prompt updater enables the learned prompt to update according to the target's motion. To capture the real-time motion of the target object, a motion information extractor is proposed to extract target-conditioned motion information between the current frame and the previous one. However, relying solely on the extraction of motion information between two consecutive frames may lead to unreliable results. This is because phenomena like occlusions and illumination changes can affect the accuracy of short-term motion information, potentially disrupting spatio-temporal continuity.

To address this issue, in addition to the short-term motion information, we also incorporate long-term motion information, thereby enhancing robustness in maintaining spatio-temporal continuity~\cite{cheng2022implicit}.

We summarize our contributions: i) From a novel perspective, we perform unsupervised object tracking, by leveraging the rich knowledge embedded in the pre-trained text-to-image diffusion model. ii) To harness the potential of the diffusion models for unsupervised object tracking, we introduce Diff-Tracker, a novel framework with crafted design components. The components include an initial prompt learner for learning an initial prompt representing the target and an online prompt updater for updating the learned prompt based on the target's motion information. iii) Our method achieves state-of-the-art performance in  unsupervised object tracking task.

\section{Related Work}
\label{sec:related}

\subsubsection{Unsupervised visual tracking.}
Thanks to deep learning techniques, significant advancements have been made in various domains of computer vision, such as object detection~\cite{hu2018relation,tan2020efficientdet,xie2021oriented}, semantic segmentation~\cite{xu2023meta,zhang2022distilling,yu2018bisenet}, image classification~\cite{zhao2023mixpro,zhao2024ltgc,zhang2024instance}, video understanding~\cite{xu2021sutd,chen2024videollm,wong2022assistq,he2021modeling}, and unsupervised visual tracking~\cite{bertinetto2016fully,zheng2021learning,shen2022unsupervised,wang2019unsupervised,sio2020s2siamfc}.
The groundbreaking deep learning-based unsupervised visual tracking work UDT~\cite{wang2019unsupervised} developes a tracker based on Discriminative Correlation Filters. This tracker is trained through forward-backward tracking of frames, under the guidance of a consistency loss function. Besides, $s^2$siamfc~\cite{sio2020s2siamfc} utilizes a Siamese pipeline, similar to SiamFC~\cite{bertinetto2016fully}, for training a foreground/background classifier with single-frame pairs. This method incorporates learning adversarial masking techniques to generate template-search pairs from identity frame. Following $s^2$siamfc, state-of-the-art unsupervised trackers~\cite{zheng2021learning,shen2022unsupervised} also employ similar siamese network structure. Differently, here we leverage the rich knowledge embedded within text-to-image diffusion models for the first time to perform unsupervised visual tracking.

\subsubsection{Text-to-image diffusion models.}
Recent developments~\cite{rombach2022high,hertz2022prompt,saharia2022photorealistic} in text-to-image diffusion models have significantly enhanced the capacity of generating diverse images from input prompts. With the advent of these pre-trained text-to-image diffusion models, \eg Stable Diffusion~\cite{rombach2022high} and Imagen~\cite{saharia2022photorealistic}, there has been a surge in research aiming to leverage powerful capabilities of these models to further advance developments in other fields, including image editing~\cite{kawar2023imagic}, text-to-video generation~\cite{khachatryan2023text2video}, domain adaptation~\cite{peng2023unsupervised}, object counting~\cite{hui2024class},  pose estimation~\cite{Gong2023,peng2024harnessing}, image inpainting~\cite{xie2023smartbrush}, human mesh recovery~\cite{Foo2023}, single-view object reconstruction~\cite{liu2023zero}, and semantic segmentation~\cite{khani2023slime}. 
The landscape of these fields is undergoing significant transformation, driven by the versatility and effectiveness of the diffusion technique. Different from their work, we extend the application of the text-to-image diffusion models to the domain of unsupervised visual tracking.

\section{Preliminaries: Text-to-Image Diffusion Models}
\label{sec:Preliminaries}
Text-to-image diffusion models are designed to reconstruct an image from a random Gaussian noise. This is achieved by progressively denoising the noise through a reverse diffusion process, which is conditioned on a text prompt \( p \). Below, we use the Stable Diffusion~\cite{rombach2022high}, a powerful diffusion model, as an example to illustrate the training process of the text-to-image diffusion models, as well as the cross-attention and self-attention mechanisms within the diffusion model.

\subsubsection{Training process.}
During the training phase, an input image $I$ and a corresponding text prompt $p$ are encoded by an image encoder $\mathcal{E}_i$ and a text encoder $\mathcal{E}_t$, respectively. Noise $\epsilon$  is added to the latent representation of the input image $\mathcal{E}_i(I)$, which is used for denoising training. The training objective of the Stable Diffusion model is to 
predict the added noise $\epsilon$ in the encoded image. This is mathematically formulated as:
\begin{equation}
\label{eq:dm}
L_{DM} = \mathbb{E}_{\mathcal{E}_i(I), \epsilon \sim \mathcal{N}(0,1), t}\left[\| \epsilon - \epsilon_{\theta}\left(z_t, t, \mathcal{E}_t(p) \right)\|_2^2\right],
\end{equation}
Where $t$ denotes the time step in the diffusion process, and $\epsilon_{\theta}$ is a denoising   UNet~\cite{ronneberger2015u} network.

\subsubsection{Cross-attention layers.}
In the Stable Diffusion, interplay between the image content and text prompt is encapsulated within the cross-attention layers of the denoising UNet. Specifically, within a cross-attention layer, the query $Q_c$ originates from the noisy image latent, while the key $K_c$ and value $V_c$ are derived from the text embedding $\mathcal{E}_t(p)$. 
The cross-attention map can be formulated as:
\begin{equation}
\label{eq:m_c}
M_c = \operatorname{Softmax}\left(\frac{Q_c K_c^T}{\sqrt{d}}\right),
\end{equation}
 where $d$ represents the projection dimension of $Q_c, K_c$, and $V_c$.
The cross-attention map, $M_c$, obtained by calculating the correlation between the image content represented in $Q_c$ and the semantic text content in $K_c$, serves to effectively highlight the image regions semantically related to the text prompt.

\subsubsection{Self-attention layers.}
In the architecture of the Stable Diffusion, along with the cross-attention layers, self-attention layers are also integrated within the UNet. These layers are instrumental in capturing semantic correlations among pixels in an image. Specifically, the self-attention map $M_s$ can be defined as:
\begin{equation}
\label{eq:m_s}
M_s = \operatorname{Softmax}\left(\frac{Q_s K_s^{T}}{\sqrt{d}}\right),
\end{equation}
which is in a same format with the cross-attention map $M_c$. The only difference is that, the query $Q_s$, the key $K_s$, and the value $V_s$ in the self-attention map $M_s$ are all derived from the noisy image latent representation. 
The self-attention map $M_s$ encapsulates information about the semantic relationships among pixels. This is because the derivation of the self-attention map involves calculating the correlation between $Q_s$ and $K_s$, both of which originate from the image pixels.

\section{Diff-Tracker}
\label{sec:method}
In this section, we first revisit the task setup of the unsupervised visual tracking task and briefly introduce framework of our Diff-Tracker~(\cref{subsec:task}), followed by a detailed description of two components (\ie the initial prompt learner~(\cref{subsec:learner}) and online prompt updater~(\cref{subsec:updater})) of the Diff-Tracker.

\subsection{Task Definition and Our Framework}
\label{subsec:task}

\subsubsection{Task definition.}
Given the location (bounding box annotation) of an arbitrary target in the first frame, unsupervised visual tracking task aims to accurately predict the locations of this target within the following video frame sequence, while the tracker is trained exclusively on unlabeled videos.

\subsubsection{Our framework.}
The purpose of designing the Diff-Tracker is to utilize the abundant knowledge embedded in the pre-trained text-to-image diffusion model to assist in unsupervised visual tracking. Inspired by~\cite{khani2023slime}, we observe that inputting an image and a text prompt into a pre-trained text-to-image diffusion model allows the model to activate regions semantically related to the text prompt on the cross-attention map of the UNet, leveraging the extensive knowledge embedded within the diffusion model. Furthermore, this prompt can accurately activate regions representing its semantics across different input images, demonstrating strong generalization capabilities. 

Based on this, we aim to learn a prompt representing the target to facilitate unsupervised visual tracking with the aid of the diffusion model. Since we do not have text and some targets cannot be explained with text, in this paper, we use prompt embedding as the prompt that needs to be learned. However, acquiring such a prompt is not straightforward. One challenge lies in encoding the relationships between the target and the background, which are crucial for effective target tracking~\cite{zhu2015weighted}. Another challenge pertains to updating the learned prompt based on the target's motion, considering that the target's appearance may undergo changes as it moves.

\begin{figure}[tb]
  \centering
  \includegraphics[width=1\textwidth]{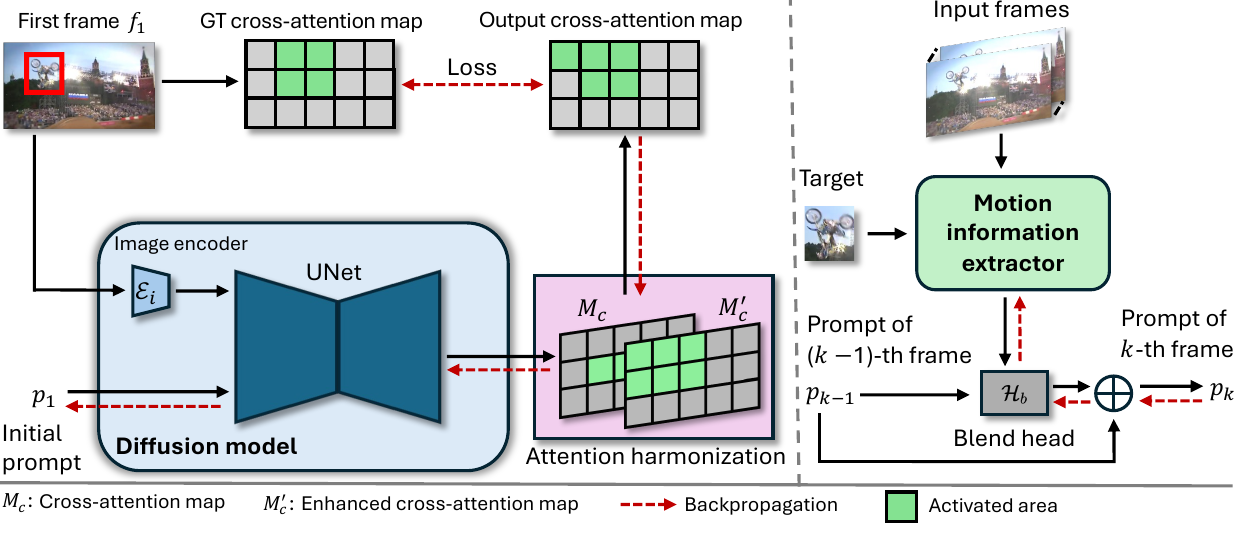}
  \caption{
  The framework of the Diff-Tracker consists of the initial prompt learner on the left side of the figure and the online prompt updater on the right. The prompt updated through the online prompt learner is input into the  network of initial prompt updater to obtain the output cross-attention map. This map is used to compute the loss for updating the online prompt updater by comparison with the GT cross-attention map.}

  \label{fig:pipeline}
\end{figure}

To address these issues, we divide the process of learning a prompt representing the target into two distinct phases. As shown in~\cref{fig:pipeline}, the first phase involves learning an initial prompt activating the region corresponding to the target object in the first frame, for which we design an initial prompt learner~(\cref{subsec:learner}). In the second phase, the learned initial prompt undergoes updates to accommodate the target's motion, ensuring continuous tracking of the target throughout the video sequence. To achieve this objective, we introduce an online prompt updater (\cref{subsec:updater}). By integrating the functionalities of the initial prompt learner and the online prompt updater, the Diff-Tracker framework is established, enabling effective and continuous tracking of the target.

\subsection{Initial Prompt Learner}
\label{subsec:learner}
The goal of the initial prompt learner is to make the text-to-image diffusion model aware of the target to be tracked. Inspired by~\cite{khani2023slime}, we achieve this goal by learning a prompt that activates the region corresponding to the target on the cross-attention map within the diffusion model. Besides, in the process of visual tracking, we may encounter common perturbations such as significant deformation or occlusion of the target, as well as the presence of nearby objects with appearances similar to the target. Under these circumstances, the relationship between the target and its background serves as crucial information for tracking~\cite{zhu2015weighted,fang2017spatial}.

To encapsulate the relationship between the target and the background in both the initial prompt and the updated prompt, we design an attention harmonization method that utilizes the self-attention mechanism of the diffusion model, because the self-attention map encodes semantic relationships between pixels. 
It is important to highlight that our attention harmonization method further allows the prompt, updated by the online prompt updater, to encapsulate the relationship between the target and its background in subsequent frames. This capability stems from the online prompt updater's utilization of the model from the initial prompt learner and its attention harmonization method.

The subsequent subsections delineate the designed initial prompt learner.

\subsubsection{Attention harmonization.}
Considering the self-attention map of the pre-trained text-to-image diffusion model is able to capture semantic relationships between pixels (as mentioned in the Section \ref{sec:Preliminaries}), we develop a harmonization method (see~\cref{fig:pipeline}) to combine the self-attention map with the cross-attention map, aiming to encode the relationship between the target and its background within the learned prompt.
Specifically, for the cross-attention map $M_c$ extracted from the UNet ( Eq. \ref{eq:m_c}), each pixel on $M_c$ can be associated with a self-attention map $M_s$ (Eq. \ref{eq:m_s}) to reflect the pixel's relationship with other pixels. We enhance the cross-attention map $M_c$ using the self-attention map $M_s$, thereby encoding the relationships among pixels. The enhanced cross-attention map $M_c^\prime$ is calculated as follows: 
\begin{equation}
M_c^\prime(:,:)=\sum_{i=1} \sum_{j=1} M_c(i, j) \cdot M_s(i, j,:,:),
\end{equation}
where $M_c(i, j)$ represents the attention value at position $(i, j)$ on the cross-attention map $M_c$, while $M_s(i, j,:,:)$ denotes the self-attention map 
related to pixel $(i, j)$ of the cross-attention map $M_c$.

Upon obtaining $M_c^\prime$, which encodes the inter-pixel relationships, we proceed to integrate $M_c^\prime$ with the original cross-attention map $M_c$ to form the final output cross-attention map $\mathcal{M}$.  To achieve this aim, we resize $M_c^\prime$ to the same size as $M_c$ and perform an element-wise  weighted summation of these two attention maps to obtain the final output cross-attention map $\mathcal{M}$. The map $\mathcal{M}$ can be defined as follows:
\begin{equation}
\mathcal{M} = (1 - \alpha) \cdot M_c^\prime + \alpha \cdot M_c,
\end{equation}
where \(\alpha\) represents a hyper parameter to balance the weights of $M_c^\prime$ and $M_c$.

We optimize the prompt embedding of the initial prompt by calculating the MSE (Mean Squared Error) loss between $\mathcal{M}$ and the GT cross-attention map. In this paper, the GT cross-attention map is defined as an attention map that only activates the bounding box area.

\subsubsection{Learning of initial prompt.} 
Finally, we describe the learning process of initial prompt $p_1$ for target object in the first frame $f_1$. Specifically, for the given $f_1$, we  firstly employ the diffusion model's image encoder $\mathcal{E}_i$ to encode $f_1$ into latent space, and get encoded representation of $f_1$. Following the workflow of the diffusion model, noise is added to this encoded representation, resulting in a noisy latent representation $z$ of $f_1$. $z$ is fed into the denoising UNet network $\epsilon_{\theta}$, accompanied by the prompt $p_1$. Subsequently, we extract the cross-attention map $M_c$ from the UNet network $\epsilon_{\theta}$ and enhance it using the self-attention map $M_s$. Both the original cross-attention map $M_c$ and the enhanced one $M_c^\prime$ are then  fused through our designed attention harmonization method to obtain the integrated cross-attention map $\mathcal{M}$. Ultimately, we calculate the normalized MSE loss between $\mathcal{M}$ and the GT cross-attention map $\mathcal{F}_1$. Overall, the loss function is defined as follows:
\begin{equation}
\label{eq:loss1}
L=\left\|\mathcal{M}-\mathcal{F}_1\right\|_2^2+ L_{D M},
\end{equation}
where  $L_{D M}$ is the loss of the diffusion model~(see the Eq. \ref{eq:dm}). The use of  loss $L_{D M}$ is to restrict the learned prompt $p_1$ resides within the text embedding space understandable by the diffusion model. It is worth noting that during the process of learning \( p_1 \), the parameters of the diffusion model are frozen.

The learned initial prompt $p_1$ is subsequently fed into the online prompt updater for online updating of the prompt.

\subsection{Online Prompt Updater}
\label{subsec:updater}
Now that we have obtained the initial prompt $p_1$ representing the target object from the first frame, we can proceed with visual tracking in the subsequent frames. In an ideal scenario, we can utilize the initial prompt to track the target throughout the subsequent frames. Yet, the object may undergo appearance changes due to prolonged motion over time, and the initial prompt  $p_1$, might fail to continuously track the object. To address this issue, we design an online prompt updater that takes into account the motion information of the target to dynamically update the prompt for the subsequent frames.

To capture the real-time motion of the target, we utilize the motion information between the current frame and the previous frame  as a basis for updating the prompt. However, due to occlusions and changes in illumination during the target's motion, such target-conditioned short-term motion between two consecutive frames 
may exhibit limited spatio-temporal coherence. Thus, to perform a more reliable prompt update process, instead of solely relying on short-term motion to update the prompt, we also incorporate the long-term motion information during the update process. This is because that the long-term motion of moving objects typically demonstrates strong spatio-temporal continuity~\cite{cheng2022implicit}.

Specifically, for each following frame, we first encode the target's motion information using a  motion information extractor. To incorporate the encoded target motion information into the updated prompt $p_k$, we design a blend head $\mathcal{H}_b$ to combine the encoded motion information with the prompt $p_{k-1}$ of the previous frame.  Finally, the output from the designed blend head $\mathcal{H}_b$ is combined with the previous frame's target prompt $p_{k-1}$ via a residual manner. This residual manner is designed to enhance the stability of our online prompt updater. 

We sequentially introduce these steps in the following subsections.

\begin{figure}[tb]
  \centering
  \includegraphics[width=1\textwidth]{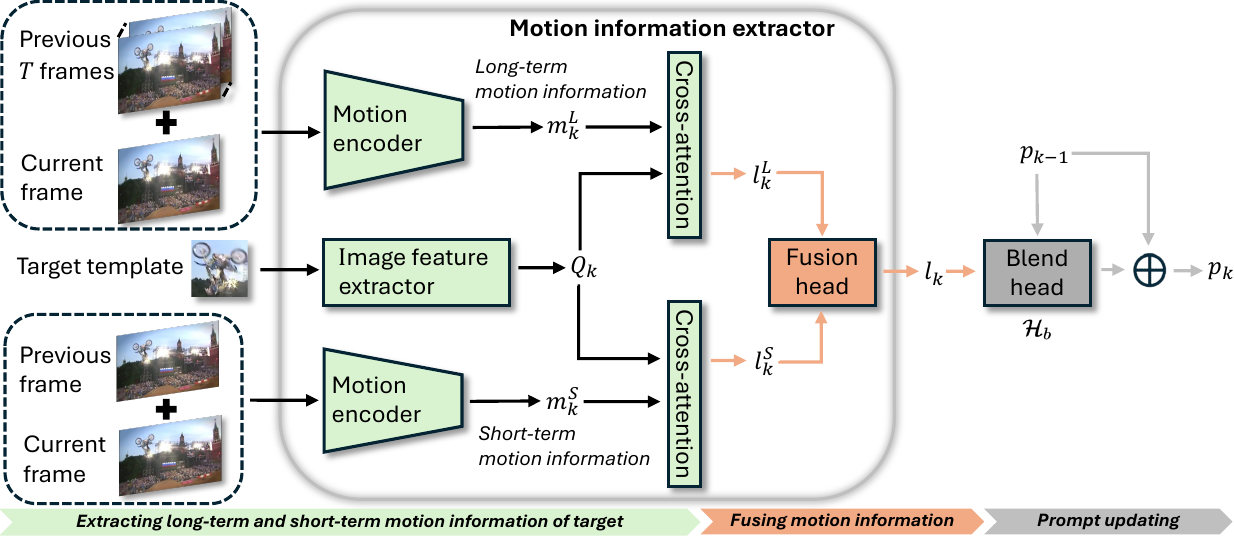}
  \caption{ The detailed architectures of the online prompt updater.
}
  
  \label{fig:updater}
\end{figure} 

\subsubsection{Motion information extractor.} To enable reliable updates of the learned prompt based on the target's motion, we first encode the long-term and short-term motion information. Specifically, as can be seen in~\cref{fig:updater}, we stack the current frame and previous $T$ frames preceding it as a video sequence and input them into one motion encoder  to obtain the long-term motion information $m_k^L$, and we input the current frame and the previous frame into another motion encoder to obtain the short-term motion information $m_k^S$.

Since the two types of encoded motion information are based  on the entire image and we only track the target, we need to derive the target-conditioned motion information (denoted as $l_k$ below) based on these two types of global motion. To this end, we utilize the cross-attention mechanism to establish connections between the target's appearance features and the two kinds of motion information. Specifically, we construct one cross-attention map by using the target's appearance features as the query, with the long-term motion information serving as the key and value. Furthermore, we construct another cross-attention map using the appearance features of the target as the query and the short-term motion information as the key and value. The two cross-attention maps are illustrated in the following equation:
\begin{equation}
l_k^L  =  \operatorname{Cross-Attn} (Q_k,m_k^L), \quad  l_k^S  =  \operatorname{Cross-Attn} (Q_k,m_k^S),  
\end{equation}
where $l_k^L$ and $l_k^S$ respectively represent the long-term and short-term motion information conditioned on the target, while $Q_k$ denotes the query. The query \(Q_k\) is obtained from the appearance features of the target, which are extracted using an image feature extractor.  We input the target template from the first frame into the image feature extractor to obtain the target's appearance features.

Then, as shown in~\cref{fig:updater}, to obtain $l_k$, we design a fusion head to merge the long-term and short-term motion information, $l_k^L$ and $l_k^S$. The fusion head is a multi-layer perceptron (MLP) network consisting of two fully connected layers and a ReLU activation layer. The obtained target's motion information $l_k$ serves as crucial information for updating the prompt.

\subsubsection{Process of prompt updating.}
As can be seen in~\cref{fig:updater}, we update the prompt representing the target in accordance with the target's motion, accommodating changes in appearance that may occur as a result of such motion. To be specific, we use a blend head $\mathcal{H}_b$ to fuse the target's motion information $l_k$, which is obtained through the motion information extractor, with the prompt of the previous frame.   Finally, prompt for the current $k$-th frame is derived by combining the output from the blend head $\mathcal{H}_b$ with the prompt of the $(k-1)$-th frame $p_{k-1}$ via a residual manner. Formally, the above prompt updating process for the $k$-th frame is defined as:
\begin{equation}
p_k  = (1 - \beta) \cdot \mathcal{H}_b(p_{k-1} + l_k) + \beta \cdot p_{k-1},  
\end{equation}
where $\beta$ is a hyper parameter to balance the weight of different terms. The loss for updating the online prompt updater is obtained by calculating the MSE loss between the cross-attention map generated by the prompt $p_k$ and the GT cross-attention map of the $k$-th frame. 

The obtained $p_k$ obtained through the online updating is used for tracking the target on the $k$-th frame.

\section{Experiments}
In this section, we first introduce the implementation details~(\cref{subsec:implementation})  of the Diff-Tracker, as well as the main experimental results~(\cref{subsec:main results}) and ablation studies~(\cref{subsec:ablation}).

\subsection{Implementation Details}
\label{subsec:implementation}

\subsubsection{Pseudo label generation.}

To obtain pseudo labels, we follow prior work~\cite{zheng2021learning} by employing an off-the-shelf optical flow model~\cite{liu2020learning} on datasets GOT-10k~\cite{huang2019got}, ImageNet VID~\cite{russakovsky2015imagenet}, LaSOT~\cite{fan2019lasot}, and YouTube-VOS~\cite{xu2018youtube}.  The sampling strategy for the pseudo labels is in accordance with~\cite{zheng2021learning}. In this paper, for fairness in comparison with previous unsupervised trackers~\cite{zheng2021learning,shen2022unsupervised}, the GT cross-attention is defined as an attention map that exclusively activates within the area of the bounding box.

\subsubsection{Model architecture.}
Our initial prompt learner is a pre-trained text-to-image diffusion model~\cite{rombach2022high}. In the online prompt updater, both two motion encoders are designed by extending a ResNet18~\cite{he2016deep} model, pre-trained on the ImageNet~\cite{deng2009imagenet} dataset, with two Conv3D layers. The image feature extractor is a also ResNet-18 model pre-trained on the ImageNet dataset. Both the fusion head and blend head are the multi-layer perceptron network consisting of two fully connected layers and a ReLU activation layer.

\subsubsection{Training.}
Our experiments are carried out using RTX 3090 GPUs. For these experiments, we set our input image size to $512\times512$, the shape of feature maps ($M_c$, $M_c^\prime$, and $\mathcal{M}$) to $1\times64\times64$.  The dimension of the prompt embedding is 1024. The hyper parameters $\alpha$ and $\beta$ are set to 0.5 and 0.7, respectively. During training, the parts that need to be learned in our Diff-Tracker are the initial prompt and the parameters in online prompt updater, and the pre-trained text-to-image diffusion model is frozen. For the training of the initial prompt, we set the learning rate  at \(5 \times 10^{-3}\) for 3 epochs with the Adam optimizer~\cite{kingma2014adam}. Besides, we train the online prompt updater for 35 epochs with the Adam optimizer and learning rate of \(5 \times 10^{-4}\).

\subsubsection{Testing.}
When testing with a given video, we first utilize the initial prompt learner with the first frame to learn the target's initial prompt. The experimental setup for learning this initial prompt is the same as the setup for learning the initial prompt during the training phase. Starting from the sixth frame, we update the learned prompt with the online prompt updater for every subsequent frame and use the updated prompt for the visual tracking. To conform to the requirements of standard tracking benchmarks, we follow~\cite{paul2022robust} to report the smallest axis-aligned bounding box that encapsulates the activated area in the cross-attention map as the bounding box.

\subsubsection{Evaluation datasets.}
We follow previous work~\cite{shen2022unsupervised} by conducting experiments on five challenging tracking datasets, including OTB2015~\cite{7001050}, VOT2016~\cite{Kristan2016}, VOT2018~\cite{Kristan2018}, TrackingNet~\cite{muller2018trackingnet} and LaSOT~\cite{fan2019lasot}.

\subsection{Main Results}
\label{subsec:main results}

\begin{table}[ht!]
\centering
\caption{We provide the evaluation results on the TrackingNet~\cite{muller2018trackingnet}, VOT2016~\cite{Kristan2016}, and VOT2018~\cite{Kristan2018} benchmark datasets. In this table, ``Unsup'' is an abbreviation for the unsupervised learning.}
\label{tab:16-18-trackingnet}
\resizebox{0.9\textwidth}{!}{
\begin{tabular}{cc|ccc|ccc|ccc}
\toprule \multirow{2}{*}{ Tracker } & \multirow{2}{*}{ Unsup } & \multicolumn{3}{|c}{ TrackingNet } & \multicolumn{3}{|c}{ VOT2016 } & \multicolumn{3}{|c}{ VOT2018 } \\
& & Suc $\uparrow$ & Pre $\uparrow$ & NPre $\uparrow$ & EAO $\uparrow$ & Acc $\uparrow$ & Rob $\downarrow$ & EAO $\uparrow$ & Acc $\uparrow$ & Rob $\downarrow$ \\
\midrule SiamFC~\cite{bertinetto2016fully} & No & 0.571 & 0.533 & 0.663 & 0.235 & 0.532 & 0.461 & 0.188 & 0.503 & 0.585 \\
DaSiamRPN~\cite{zhu2018distractor} & No & - & - & - & 0.411 & 0.610 & 0.220 & 0.326 & 0.560 & 0.340 \\
SiamRPN++~\cite{li2019siamrpn++} & No & 0.733 & 0.694 & 0.800 & - & - & - & 0.414 & 0.600 & 0.234 \\
ATOM~\cite{danelljan2019atom} & No & 0.703 & 0.648 & 0.711 & - & - & - & 0.401 & 0.590 & 0.204 \\
DiMP~\cite{bhat2019learning} & No & 0.740 & 0.687 & 0.801 & - & - & - & 0.440 & 0.597 & 0.153 \\
\midrule KCF~\cite{henriques2014high} & Yes & 0.447 & 0.419 & 0.546 & 0.192 & 0.489 & 0.569 & 0.135 & 0.447 & 0.773 \\
ECO~\cite{danelljan2017eco} & Yes & 0.561 & 0.489 & 0.621 & 0.375 & 0.550 & 0.569 & 0.280 & 0.270 & 0.480 \\
S2SiamFC~\cite{sio2020s2siamfc} & Yes & - & - & - & 0.215 & 0.493 & 0.639 & 0.180 & 0.463 & 0.782 \\
LUDT+~\cite{wang2021unsupervised} & Yes & 0.563 & 0.495 & 0.633 & 0.299 & 0.570 & 0.331 & 0.230 & 0.490 & 0.412 \\
USOT~\cite{zheng2021learning} & Yes & 0.599 & 0.551 & 0.682 & 0.351 & 0.593 & 0.336 & 0.290 & 0.564 & 0.435 \\
USOT*~\cite{zheng2021learning} & Yes & 0.616 & 0.566 & 0.691 & 0.402 & 0.600 & 0.233 & 0.344 & 0.578 & 0.304 \\
 ULAST*-off~\cite{shen2022unsupervised} & Yes & 0.649 & 0.585 & 0.725 & 0.397 & 0.599 & 0.224 & 0.347 & 0.569 & 0.304 \\
ULAST*-on~\cite{shen2022unsupervised} & Yes & 0.654 & 0.592 & 0.732 & 0.417 & 0.603 & 0.214 & 0.355 & 0.571 & 0.286 \\
\midrule
Ours &Yes&0.675&0.614&0.751&0.430&0.605&0.206&0.365&0.580&0.273\\
\bottomrule
\end{tabular}
}
\end{table}

\subsubsection{VOT2016.}
The VOT2016 benchmark dataset comprises 60 video sequences. In this dataset, tracking performance is evaluated using three key metrics: Robustness (Rob), Accuracy (Acc), and Expected Average Overlap (EAO) as documented in~\cite{Kristan2016}. As shown in~\cref{tab:16-18-trackingnet}, our proposed Diff-Tracker consistently improves upon all reported metrics as compared to other unsupervised approaches~\cite{zheng2021learning,shen2022unsupervised}. Moreover, compared to supervised methods~\cite{bertinetto2016fully,danelljan2019atom}, our method still obtains competitive results.

\subsubsection{TrackingNet.}
TrackingNet represents a comprehensive large-scale benchmark tailored for evaluating tracking performance in unconstrained environments, comprising more than 30,000 video sequences. Beyond the precision and success metrics, TrackingNet introduces an additional metric known as normalized precision (NPre). Specifically, it includes a designated test set of 511 videos. As indicated in~\cref{tab:16-18-trackingnet}, the performance of our Diff-Tracker surpasses all other unsupervised trackers.

\subsubsection{VOT2018.}
The testing sequences in VOT2018 present increased challenges in comparison to those in VOT2016. The evaluation metrics used to assess the tracker's performance on this dataset are the same as those used in VOT2016. Our proposed method is compared against representative supervised and unsupervised trackers, with the evaluation results displayed in~\cref{tab:16-18-trackingnet}. Our Diff-tracker achieves the highest tracking performance among the evaluated unsupervised trackers, according to the three evaluation metrics.

\begin{table}[ht!]

\caption{We provide the evaluation results on the OTB2015~\cite{7001050} and LaSOT~\cite{fan2019lasot} benchmark datasets.}
\label{tab:otb15-lasot}
\centering
\resizebox{0.55\textwidth}{!}{
\begin{tabular}{cc|cc|cc}
\toprule \multirow{2}{*}{ Tracker } & \multirow{2}{*}{ Unsup } & \multicolumn{2}{|c}{ OTB2015 } & \multicolumn{2}{|c}{ LaSOT } \\
& & Suc $\uparrow$ & Pre $\uparrow$ & Suc $\uparrow$ & Pre $\uparrow$ \\
\midrule SiamFC~\cite{bertinetto2016fully}  & No & 0.582 & 0.771 & 0.336 & 0.339 \\
SiamRPN~\cite{li2018high}  & No & 0.637 & 0.851 & 0.411 & 0.380 \\
SiamRPN++~\cite{li2019siamrpn++}  & No & 0.696 & 0.923 & 0.495 & 0.493 \\
\midrule KCF~\cite{henriques2014high}  & Yes & 0.485 & 0.696 & 0.178 & 0.166 \\
DSST~\cite{danelljan2014accurate}  & Yes & 0.518 & 0.689 & 0.207 & 0.189 \\
LUDT+~\cite{wang2021unsupervised}  & Yes & 0.639 & 0.843 & 0.305 & 0.288 \\
USOT~\cite{zheng2021learning}  & Yes & 0.589 & 0.806 & 0.337 & 0.323 \\
USOT*~\cite{zheng2021learning}  & Yes & 0.574 & 0.775 & 0.358 & 0.340 \\
ULAST*-off~\cite{shen2022unsupervised} & Yes & 0.645 & 0.878 & 0.468 & 0.448 \\
ULAST*-on~\cite{shen2022unsupervised} & Yes & 0.648 & 0.879 & 0.471 & 0.451 \\ \midrule
 Ours &Yes & 0.661 & 0.898 & 0.486 & 0.472 \\
\bottomrule
\end{tabular}
}
\end{table}

\subsubsection{OTB2015.}
The OTB2015 dataset encompasses 100 video sequences, featuring a diverse array of targets. The evaluation metrics in the OTB2015 benchmark are the success score and precision score. The experimental results are presented in~\cref{tab:otb15-lasot}. We surpass prior unsupervised approaches across all metrics, demonstrating the effectiveness of our Diff-Tracker.

\subsubsection{LaSOT.}
LaSOT is an extensive annotated collections within the tracking community, comprising 280 extended video sequences. The evaluation metric for this benchmark aligns with that of the OTB2015. We achieve the state-of-the-art performance on the LaSOT dataset, as presented in~\cref{tab:otb15-lasot}. For more detailed analysis, please refer to our supplementary.

\subsection{Ablation Study}
\label{subsec:ablation}

\begin{table}[ht!]
 \caption{We evaluate the impact of the attention harmonization method and online prompt updater, utilizing the VOT2018~\cite{Kristan2018} benchmark dataset. In the table, ``Ours (w/o Attention Harmonization Method)'' refers to our method without the designed attention harmonization method.}
\label{tab:ablation}
\centering
\resizebox{0.7\textwidth}{!}{
\begin{tabular}{c|ccc}
\toprule     & EAO $\uparrow$  & Acc $\uparrow$ & Rob $\downarrow$  \\
\midrule 
Ours (w/o attention harmonization method) &  0.359 & 0.577 & 0.280 \\
Ours (w/o online prompt updater) &    0.349 & 0.571 & 0.292\\
\midrule 
Ours     &0.365 &0.580 &0.273 \\
\bottomrule
\end{tabular}
}
\end{table}

\subsubsection{Attention harmonization.}
 The impact of the harmonization of the self-attention map and the cross-attention map  is also explored. Without using the attention harmonization method, we directly compute the MSE loss between the cross-attention map $M_c$, which is extracted from UNet network of the diffusion model, and the generated GT cross-attention map. As shown in~\cref{tab:ablation}, the addition of the attention harmonization method leads to a performance improvement, showing its efficacy.

\subsubsection{Online prompt  updater.}
As shown in~\cref{tab:ablation}, we also verify the impact of the designed online prompt updater in our method. Without the online prompt updater, our Diff-Tracker directly utilizes the learned initial prompt tracking for target tracking in the following frames. From this table, we can observe that the performance of our method decreases in the absence of the online prompt  updater, demonstrating that online prompt  updater is effective in utilizing the motion information to update the learned prompt, thereby better representing the target.

\begin{table}[h!]

 \caption{We evaluate the impact of the long-term motion information and short-term motion information, utilizing the VOT2018~\cite{Kristan2018} benchmark dataset. In the table, ``Ours (w/o long-term motion information)'' refers to our online prompt updater without the long-term motion information. }
\label{tab:motion}
\centering
\resizebox{0.7\textwidth}{!}{
\begin{tabular}{c|ccc}

\toprule      & EAO $\uparrow$  & Acc $\uparrow$ & Rob $\downarrow$  \\
\midrule 
Ours (w/o long-term motion information)  &0.355  & 0.572  & 0.286  \\
  Ours (w/o short-term motion information)  & 0.360 & 0.577  & 0.281  \\
\midrule 
Ours &0.365 &0.580 &0.273  \\
\bottomrule
\end{tabular}
}

\end{table}

\subsubsection{Long-term and short-term motion information.}
We explore the impact of long-term and short-term motion information (see~\cref{tab:motion}). The experimental results show that our method performs best when both long-term and short-term motion information are utilized, which validates the effectiveness of combining these two types of motion information for online updating of the prompt representing the target. For more detailed analysis, please refer to our supplementary.

\section{Conclusion}
To tackle the challenging task of the unsupervised visual tracking, we develop Diff-Tracker, which leverages the powerful capabilities of the pre-trained text-to-image diffusion models. The Diff-Tracker learns an initial prompt representing the tracking target through a designed initial prompt learner and updates the learned prompt based on the target's motion using a proposed online prompt updater. Through extensive experiments, we observe that our Diff-Tracker outperforms state-of-the-art unsupervised trackers across five widely used visual tracking benchmarks.

\section*{Acknowledgements}
This research is supported by the Ministry of Education, Singapore, under the AcRF Tier 2 Projects (MOE-T2EP20222-0009 and MOE-T2EP20123-0014), and the National Research Foundation Singapore through its AI Singapore Programme (AISG-100E-2023-121).

\bibliographystyle{splncs04}
\bibliography{main}
\end{document}